\documentclass[11pt,letterpaper]{article}
\usepackage{emnlp2017}
\usepackage{times}
\usepackage{latexsym}
\usepackage{amsmath}
\usepackage{amssymb}
\usepackage{graphicx} 
\usepackage{tikz}
\usepackage{tikz-qtree}
\usepackage{subcaption}
\usepackage{caption}

\emnlpfinalcopy

\def\emnlppaperid{7}

\newcommand{\RR}{\mathbb{R}}

\title{Recurrent Neural Network-Based Sentence Encoder with Gated Attention for Natural Language Inference}

\author{
Qian Chen \\
University of Science and  
Technology of China\\
\tt{cq1231@mail.ustc.edu.cn} \\\And
Xiaodan Zhu \\
Queen's University \\
\tt{xiaodan.zhu@queensu.ca} \\\AND
Zhen-Hua Ling \\
University of Science and
Technology of China\\
\tt{zhling@ustc.edu.cn} \\\And
Si Wei \\
iFLYTEK Research\\
\tt{siwei@iflytek.com} \\\AND
Hui Jiang \\
York University\\
\tt{hj@cse.yorku.ca} \\\And 
Diana Inkpen \\
University of Ottawa\\
\tt{diana@site.uottawa.ca}
}

\date{}

\begin{document}

\maketitle

\begin{abstract}
The RepEval 2017 Shared Task aims to evaluate natural language understanding models for sentence representation, in which a sentence is represented as a fixed-length vector with neural networks and the quality of the representation is tested with a natural language inference task. This paper describes our system (alpha) that is ranked among the top in the Shared Task, on both the in-domain test set (obtaining a 74.9\% accuracy) and on the cross-domain test set (also attaining a 74.9\% accuracy), demonstrating that the model generalizes well to the cross-domain data. Our model is equipped with intra-sentence gated-attention composition which helps achieve a better performance. In addition to submitting our model to the Shared Task, we have also tested it on the Stanford Natural Language Inference (SNLI) dataset. We obtain an accuracy of 85.5\%, which is the best reported result on SNLI when cross-sentence attention is not allowed, the same condition enforced in RepEval 2017.

\end{abstract}

\section{Introduction}
The RepEval 2017 Shared Task aims to evaluate language understanding models for sentence representation with natural language inference (NLI) tasks, where a sentence is represented as a fixed-length vector. 

Modeling inference in human language is very challenging but is a basic problem in natural language understanding. Specifically, NLI is concerned with determining whether a hypothesis sentence~\textit{h} can be inferred from a premise sentence~\textit{p}.

Most previous top-performing neural network models on NLI use attention models between a premise and its hypothesis, while how much information can be encoded in a fixed-length vector without such cross-sentence attention deserves some further understanding. In this paper, we describe the model we submitted to the RepEval 2017 Shared Task~\citep{nangia2017repeval}, which achieves the top performance on both the in-domain and cross-domain test set. 

\section{Related Work}
Natural language inference (NLI), also named recognizing textual entailment (RTE) includes a large bulk of early work on rather small datasets with more conventional methods~\citep{Dagan2005ThePR,MacCartneyThesis}. More recently, the large datasets are available, which makes it possible to train natural language inference models based on neural networks~\citep{Bowman:D15-1075,DBLP:journals/corr/WilliamsNB17}. 

Natural language inference models based on neural networks are mainly separated into two kind of ways, sentence encoder-based models and cross-sentence attention-based models. Among them, Enhanced Sequential Inference Model (ESIM) with cross-sentence attention represents the state of the art~\citep{DBLP:journals/corr/ChenZLWJ16}. However, in this paper we principally concentrate on sentence encoder-based model. Many researchers have studied sentence encoder-based model for natural language inference~\citep{Bowman:D15-1075,DBLP:journals/corr/VendrovKFU15,Mou:P16-2022,Bowman:P16-1139,DBLP:journals/corr/MunkhdalaiY16,DBLP:journals/corr/MunkhdalaiY16b,DBLP:journals/corr/LiuSLW16,DBLP:journals/corr/LinFSYXZB17}. It is, however, not very clear if the potential of the sentence encoder-based model has been well exploited. In this paper, we demonstrate that proposed models based on gated-attention can achieve a new state-of-the-art performance for natural language inference. 

\section{Methods}

\begin{figure}[!htb]
	\centering
	\includegraphics[width=\linewidth]{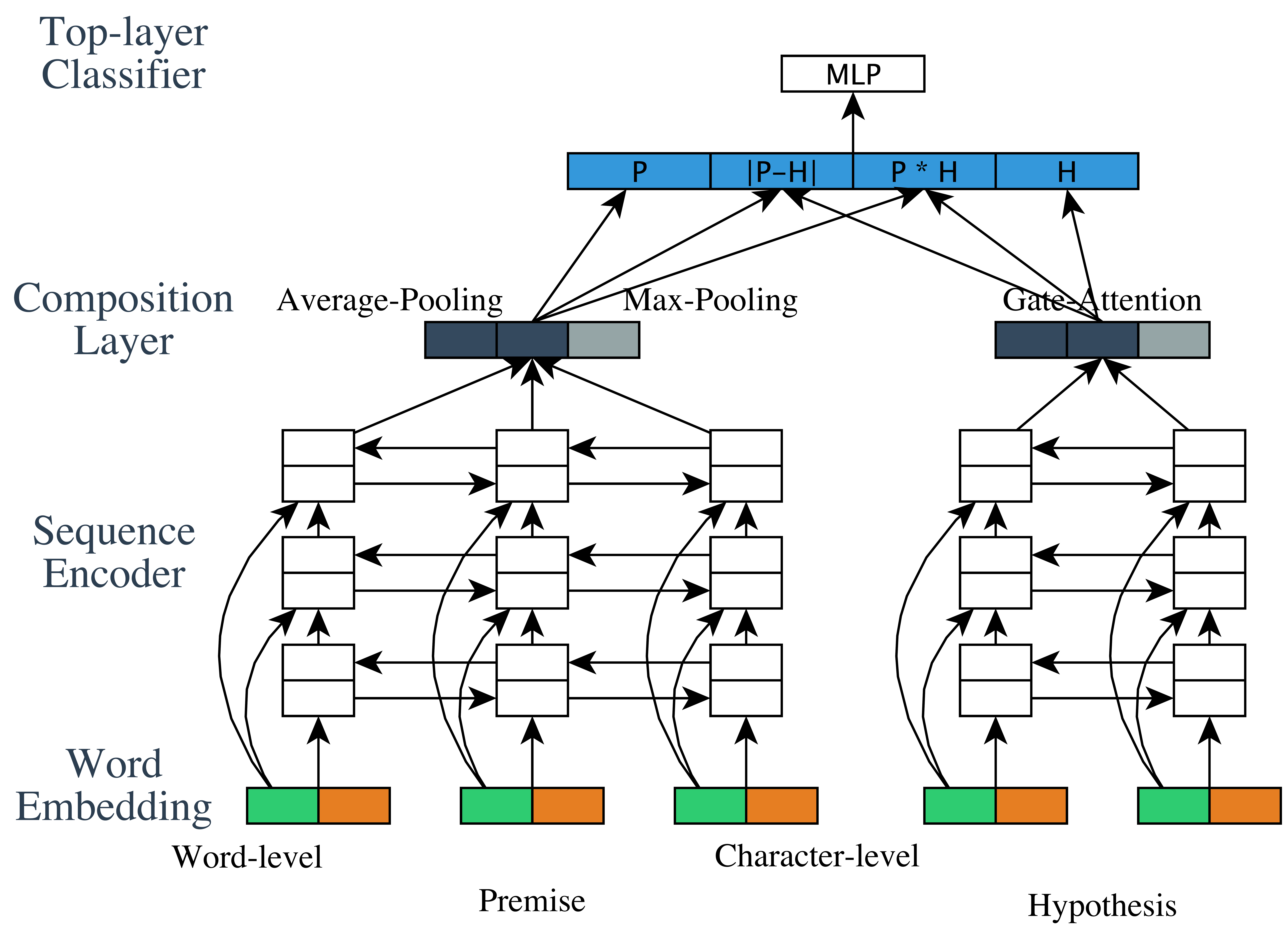}
	\caption{A view of our neural language inference network.}
    \vspace{-2mm}
	\label{fig:nli}
\end{figure}

We present here the proposed natural language inference networks which are composed of the following major components: word embedding, sequence encoder, composition layer, and the top-layer classifier. 
Figure~\ref{fig:nli} shows a view of the architecture of our neural language inference network. 

\subsection{Word Embedding}
In our notation, a sentence (premise or hypothesis) is indicated as $x=(x_1,\dots,x_l)$, where $l$ is the length of the sentence. We concatenate embeddings learned at two different levels to represent each word in the sentence: the character composition and holistic word-level embedding. The character composition feeds all characters of each word into a convolutional neural network (CNN) with max-pooling~\citep{kim2014convolutional} to obtain representations $c = (c_1,\dots,c_l)$. In addition, we also use the pre-trained GloVe vectors~\citep{Pennington:D14-1162} for each word as holistic word-level embedding $w=(w_1,\dots,w_l)$. Therefore, each word is represented as a concatenation of the character-composition vector and word-level embedding $e=([c_1;w_1],\dots,[c_l;w_l])$. This is performed on both the premise and hypothesis, resulting into two matrices: the $e^p \in \mathbb{R} ^{n \times d_w}$ for a premise and the $e^h \in \mathbb{R} ^{m \times d_w}$ for a hypothesis, where $n$ and $m$ are the length of the premise and hypothesis respectively, and $d_w$ is the embedding dimension.

\subsection{Sequence Encoder}
To represent words and their context in a premise and hypothesis, sentence pairs are fed into sentence encoders to obtain hidden vectors ($h^p$ and $h^h$). We use stacked bidirectional LSTMs (BiLSTM) as the encoders. Shortcut connections are applied, which concatenate word embeddings and input hidden states at each layer in the stacked BiLSTM except for the bottom layer.
\begin{align}
h^p&=\text{BiLSTM}({e^p}) \in \mathbb{R} ^{n \times 2d} \\
h^h&=\text{BiLSTM}({e^h}) \in \mathbb{R} ^{m \times 2d}
\end{align}

\noindent where $d$ is the dimension of hidden states of LSTMs.
A BiLSTM concatenate a forward and backward LSTM on a sequence $h_t=[\overrightarrow{h_t};\overleftarrow{h_t}]$, starting from the left and the right end, respectively. Hidden states of unidirectional LSTM ($\overrightarrow{h_t}$ or $\overleftarrow{h_t}$) are calculated as follows, 
\begin{align}
 \begin{bmatrix}
  i_t \\
  f_t \\
  u_t \\
  o_t
 \end{bmatrix}
&= 
 \begin{bmatrix}
  \sigma \\
  \sigma \\
  \tanh \\
  \sigma
  \end{bmatrix} 
 (W x_t + Uh_{t-1} + b) \\
c_t &= f_t \odot c_{t-1} + i_t \odot u_t \\
h_t &= o_t \odot \tanh(c_t)
\end{align}

\noindent  where $\sigma$ is the sigmoid function, $\odot$ is the element-wise multiplication of two vectors, and ${W} \in \RR^{4d \times d_w}$, ${U} \in \RR^{4d \times d}$, $b \in \RR^{4d \times 1}$ are weight matrices to be learned. For each input vector $x_t$ at time step $t$, LSTM applies a set of gating functions---the input gate $i_t$, forget gate $f_t$, and output gate $o_t$, together with a memory cell $c_t$, to control message flow and track long-distance information~\citep{DBLP:journals/neco/HochreiterS97} and generate a hidden state ${h}_t$ at each time step. 

\subsection{Composition Layer}
To transform sentences into fixed-length vector representations and reason using those representations, we need to compose the hidden vectors obtained by the sequence encoder layer ($h^p$ and $h^h$). We propose intra-sentence gated-attention to obtain a fixed-length vector. Illustrated by the case of hidden states of premise $h^p$,

\begin{align}
\label{equ:gate}
v^p_g =& \sum_{t=1}^n \frac{\lVert i_t \rVert_2}{\sum_{j=1}^n{\lVert i_j \rVert_2}} h^p_t \\ 
\text{or}~v^p_g =& \sum_{t=1}^n \frac{\lVert 1-f_t \rVert_2}{\sum_{j=1}^n{\lVert 1-f_j \rVert_2}} h^p_t \\
\text{or}~v^p_g =& \sum_{t=1}^n \frac{\lVert o_t\rVert_2}{\sum_{j=1}^n{\lVert o_j \rVert_2}} h^p_t 
\end{align}

\noindent where $i_t$, $f_t$, $o_t$ are the input gate, forget gate, and output gate in the BiLSTM of the top layer. Note that the gates are concatenated by forward and backward LSTM, i.e., $i_t=[\overrightarrow{i_t};\overleftarrow{i_t}]$, $f_t=[\overrightarrow{f_t};\overleftarrow{f_t}]$, $o_t=[\overrightarrow{o_t};\overleftarrow{o_t}]$. $\lVert*\rVert_2$ indicates $l^2$-norm, which converts vectors to scalars. The idea of gated-attention is inspired by the fact that human only remember important parts after they read sentences.~\citep{DBLP:journals/corr/LiuSLW16,DBLP:journals/corr/LinFSYXZB17} proposed a similar ``inner-attention'' mechanism but it's calculated by an extra MLP layer which would require more computation than us.

We also use average-pooling and max-pooling to obtain fixed-length vectors $v_a$ and $v_m$ as in~\citet{DBLP:journals/corr/ChenZLWJ16}. Then, the final fixed-length vector representation of premise is $v^p = [v^p_g;v^p_a;v^p_m]$. As for hidden states of hypothesis $h^h$, we can obtain $v^h$ through similar calculation procedure. Consequently, both the premise and hypothesis are fed into the composition layer to obtain fixed-length vector representations respectively ($v^p, v^h$).

\subsection{Top-layer Classifier}
Our inference model feeds the resulting vectors obtained above to the final classifier to determine the overall inference relationship. In our models, we compute the absolute difference and the element-wise product for the tuple $[v^p, v^h]$. 

The absolute difference and element-wise product are then concatenated with the original vectors $v^p$ and $v^h$~\citep{Mou:P16-2022}. 
\begin{equation}
v_{\text{inp}} = [v^p; v^h; \lvert v^p-v^h \rvert ; v^p \odot v^h]
\end{equation}

We then put the vector $v_{\text{inp}}$ into a final multilayer perceptron (MLP) classifier. The MLP has 2 hidden layers with \textit{ReLu} activation with shortcut connections and a \textit{softmax} output layer in our experiments. The entire model (all four components described above) is trained end-to-end, and the cross-entropy loss of the training set is minimized. 

\section{Experimental Setup}

\paragraph{Data}
RepEval 2017 use Multi-Genre NLI corpus (MultiNLI)~\citep{DBLP:journals/corr/WilliamsNB17}, which focuses on three basic relationships between a premise and a potential hypothesis: the premise entails the hypothesis (\textit{entailment}), they contradict each other (\textit{contradiction}), or they are not related (\textit{neutral}). The corpus has ten genres, such as fiction, letters, telephone speech and so on. Training set only has five genres of them, therefore there are in-domain and cross-domain development/test sets. SNLI~\citep{Bowman:D15-1075} corpus can be used as an additional training/development set, which includes content from the single genre of image captions. However, we don't use SNLI as an additional training/development data in our experiments.

\paragraph{Training}
We use the in-domain development set to select models for testing. To help replicate our results, we publish our code at \url{https://github.com/lukecq1231/enc_nli} (the core code is also used or adapted for a summarization~\citep{DBLP:conf/ijcai/ChenZLWJ16} and a question-answering task~\citep{Zhang:qa:2017}). We use the Adam~\citep{DBLP:journals/corr/KingmaB14} for optimization. Stacked BiLSTM has 3 layers, and all hidden states of BiLSTMs and MLP have 600 dimensions. The character embedding has 15 dimensions, and CNN filters length is [1,3,5], each of those is 100 dimensions. We use pre-trained \textit{GloVe-840B-300D} vectors~\citep{Pennington:D14-1162} as our word-level embeddings and fix these embeddings during the training process. Out-of-vocabulary (OOV) words are initialized randomly with Gaussian samples. 

\section{Results}

\begin{table}[t!]
\renewcommand{\arraystretch}{0.9}
\centering
\begin{tabular}{|l|c|c|}
\hline
Model     & In & Cross\\
\hline
CBOW & 64.8 & 64.5 \\
BiLSTM &  66.9 & 66.9\\
ESIM & 72.3 & 72.1\\
\hline
TALP-UPC$^*$ & 67.9 & 68.2 \\
LCT-MALTA$^*$& 70.7 & 70.8 \\
Rivercorners$^*$& 72.1 & 72.1\\
Rivercorners (ensemble)$^*$& 72.2 & 72.8 \\
YixinNie-UNC-NLP$^*$ & 74.5 & 73.5 \\
\hline
Our ESIM & 76.8 & 75.8 \\
Single$^*$ & 73.5 & 73.6 \\
Ensembled$^*$ & 74.9 & 74.9 \\
\hline
Single (Input Gate)$^*$ & 73.5 & 73.6 \\
Single (Forget Gate) & 72.9 & 73.1 \\
Single (Output Gate) & 73.7 & 73.4 \\
\hline
Single - Gated-Att & 72.8 & 73.6 \\
Single - CharCNN & 72.9 & 73.5 \\
Single - Word Embedding & 65.6 & 66.0 \\
Single - AbsDiff/Product & 69.7 & 69.2 \\
\hline
\end{tabular}
\caption{Accuracies of the models on MultiNLI. Note that $^*$ indicates that the model participate in the competition on June 15th, 2017.}
\label{tab:result}
\end{table}

Table~\ref{tab:result} shows the results of different models. The first group of models are copied from~\citet{DBLP:journals/corr/WilliamsNB17}. The first sentence encoder is based on continuous bag of words (CBOW), the second is based on BiLSTM, and the third model is Enhanced Sequential Inference Model (ESIM)~\citep{DBLP:journals/corr/ChenZLWJ16}
reimplemented by~\citet{DBLP:journals/corr/WilliamsNB17}, which represents the state of the art on SNLI dataset. However, ESIM uses attention between sentence pairs, which is not a sentence-encoder based model.

The second group of models are the results of other teams which participate the RepEval 2017 Share Task competition~\citep{nangia2017repeval}. 

In addition, we also use our implementation of ESIM, which achieves an accuracy of 76.8\% in the in-domain test set, and 75.8\% in the cross-domain test set, which presents the state-of-the-art results. 
After removing the cross-sentence attention and adding our gated-attention model, we achieve accuracies of 73.5\% and 73.6\%, which ranks first in the cross-domain test set and ranks second in the in-domain test set among the single models. 

When ensembling our models, we obtain accuracies 74.9\% and 74.9\%, which ranks first in both test sets. Our ensembling is performed by averaging the five models trained with different parameter initialization.

We compare the performance of using different gate in gate-attention in the fourth group of Table~\ref{tab:result}. Note that we use attention based on input gate on all other experiments.

To understand the importance of the different elements of the proposed model, we remove some crucial elements from our single model. 
If we remove the gated-attention, the accuracies drop to 72.8\% and 73.6\%. 
If we remove character-composition vector, the accuracies drop to 72.9\% and 73.5\%. If we remove word-level embedding, the accuracies drop to 65.6\% and 66.0\%. If we remove absolute difference and element-wise product of the sentence representation vectors, the accuracies drop to 69.7\% and 69.2\%. 

\begin{table}[t!]
\renewcommand{\arraystretch}{0.9}
\centering
\begin{tabular}{|l|c|}
\hline
Model     & Test\\
\hline
LSTM~\citep{Bowman:D15-1075} & 80.6 \\
GRU~\citep{DBLP:journals/corr/VendrovKFU15} & 81.4 \\
Tree CNN~\citep{Mou:P16-2022} & 82.1 \\
SPINN-PI~\citep{Bowman:P16-1139} & 83.2 \\
NTI~\citep{DBLP:journals/corr/MunkhdalaiY16b} & 83.4 \\
Intra-Att BiLSTM~\citep{DBLP:journals/corr/LiuSLW16} & 84.2 \\
Self-Att BiLSTM~\citep{DBLP:journals/corr/LinFSYXZB17} & 84.2 \\
NSE~\citep{DBLP:journals/corr/MunkhdalaiY16} & 84.6 \\
\hline
Gated-Att BiLSTM & 85.5 \\
\hline
\end{tabular}
\caption{Accuracies of the models on SNLI. }
\label{tab:snli}
\end{table}

In addition to testing on this shared task, we have also applied our best single system (without ensembling) on the SNLI dataset; our model achieve an accuracy of 85.5\%, which is best result reported on SNLI, outperforming all previous models when cross-sentence attention is not allowed. The previous state-of-the-art sentence encoder-based model ~\citep{DBLP:journals/corr/MunkhdalaiY16b}, called neural semantic encoders (NSE), only achieved an accuracy of 84.6\% on SNLI. Table~\ref{tab:snli} shows the results of previous models and proposed model. 

\section{Summary and Future Work}

We describe our system that encodes a sentence to a fixed-length vector for natural language inference, which achieves the top performances on both the RepEval-2017 and the SNLI dataset. The model is equipped with a novel intra-sentence gated-attention component. The model only uses a common stacked BiLSTM as the building block together with the intra-sentence gated-attention in order to compose the fixed-length representations. Our model could be used on other sentence encoding tasks. Future work on NLI includes exploring the usefulness of external resources such as WordNet and contrasting-meaning embedding~\citep{Chen:P15-1011}.

\section*{Acknowledgments}
The first and the third author of this paper were supported in part by the National Natural Science Foundation of China (Grants
No. U1636201) and the Fundamental Research Funds for the Central Universities (Grant No. WK2350000001).

\clearpage
\bibliography{emnlp2017}

\begin{thebibliography}{}
\expandafter\ifx\csname natexlab\endcsname\relax\def\natexlab#1{#1}\fi

\bibitem[{Bowman et~al.(2015)Bowman, Angeli, Potts, and
  Manning}]{Bowman:D15-1075}
R.~Samuel Bowman, Gabor Angeli, Christopher Potts, and D.~Christopher Manning.
  2015.
\newblock \href{https://doi.org/10.18653/v1/D15-1075}{A large annotated corpus
  for learning natural language inference}.
\newblock In {\em Proceedings of the 2015 Conference on Empirical Methods in
  Natural Language Processing\/}. Association for Computational Linguistics,
  pages 632--642.
\newblock
  \href{https://doi.org/10.18653/v1/D15-1075}{https://doi.org/10.18653/v1/D15-1075}.

\bibitem[{Bowman et~al.(2016)Bowman, Gauthier, Rastogi, Gupta, Manning, and
  Potts}]{Bowman:P16-1139}
R.~Samuel Bowman, Jon Gauthier, Abhinav Rastogi, Raghav Gupta, D.~Christopher
  Manning, and Christopher Potts. 2016.
\newblock \href{https://doi.org/10.18653/v1/P16-1139}{A fast unified model for
  parsing and sentence understanding}.
\newblock In {\em Proceedings of the 54th Annual Meeting of the Association for
  Computational Linguistics (Volume 1: Long Papers)\/}. Association for
  Computational Linguistics, pages 1466--1477.
\newblock
  \href{https://doi.org/10.18653/v1/P16-1139}{https://doi.org/10.18653/v1/P16-1139}.

\bibitem[{Chen et~al.(2016{\natexlab{a}})Chen, Zhu, Ling, Wei, and
  Jiang}]{DBLP:conf/ijcai/ChenZLWJ16}
Qian Chen, Xiaodan Zhu, Zhen{-}Hua Ling, Si~Wei, and Hui Jiang.
  2016{\natexlab{a}}.
\newblock \href{http://www.ijcai.org/Abstract/16/391}{Distraction-based neural
  networks for modeling document}.
\newblock In Subbarao Kambhampati, editor, {\em Proceedings of the Twenty-Fifth
  International Joint Conference on Artificial Intelligence, {IJCAI} 2016, New
  York, NY, USA, 9-15 July 2016\/}. {IJCAI/AAAI} Press, pages 2754--2760.
\newblock
  \href{http://www.ijcai.org/Abstract/16/391}{http://www.ijcai.org/Abstract/16/391}.

\bibitem[{Chen et~al.(2016{\natexlab{b}})Chen, Zhu, Ling, Wei, and
  Jiang}]{DBLP:journals/corr/ChenZLWJ16}
Qian Chen, Xiaodan Zhu, Zhen{-}Hua Ling, Si~Wei, and Hui Jiang.
  2016{\natexlab{b}}.
\newblock \href{http://arxiv.org/abs/1609.06038}{Enhanced {LSTM} for natural
  language inference}.
\newblock {\em CoRR\/} abs/1609.06038.
\newblock
  \href{http://arxiv.org/abs/1609.06038}{http://arxiv.org/abs/1609.06038}.

\bibitem[{Chen et~al.(2015)Chen, Lin, Chen, Chen, Wei, Jiang, and
  Zhu}]{Chen:P15-1011}
Zhigang Chen, Wei Lin, Qian Chen, Xiaoping Chen, Si~Wei, Hui Jiang, and Xiaodan
  Zhu. 2015.
\newblock \href{https://doi.org/10.3115/v1/P15-1011}{Revisiting word embedding
  for contrasting meaning}.
\newblock In {\em Proceedings of the 53rd Annual Meeting of the Association for
  Computational Linguistics and the 7th International Joint Conference on
  Natural Language Processing (Volume 1: Long Papers)\/}. Association for
  Computational Linguistics, pages 106--115.
\newblock
  \href{https://doi.org/10.3115/v1/P15-1011}{https://doi.org/10.3115/v1/P15-1011}.

\bibitem[{Dagan et~al.(2005)Dagan, Glickman, and Magnini}]{Dagan2005ThePR}
Ido Dagan, Oren Glickman, and Bernardo Magnini. 2005.
\newblock The pascal recognising textual entailment challenge.
\newblock In {\em MLCW\/}.

\bibitem[{Hochreiter and Schmidhuber(1997)}]{DBLP:journals/neco/HochreiterS97}
Sepp Hochreiter and J{\"{u}}rgen Schmidhuber. 1997.
\newblock \href{https://doi.org/10.1162/neco.1997.9.8.1735}{Long short-term
  memory}.
\newblock {\em Neural Computation\/} 9(8):1735--1780.
\newblock
  \href{https://doi.org/10.1162/neco.1997.9.8.1735}{https://doi.org/10.1162/neco.1997.9.8.1735}.

\bibitem[{Kim(2014)}]{kim2014convolutional}
Yoon Kim. 2014.
\newblock Convolutional neural networks for sentence classification.
\newblock {\em arXiv preprint arXiv:1408.5882\/} .

\bibitem[{Kingma and Ba(2014)}]{DBLP:journals/corr/KingmaB14}
Diederik~P. Kingma and Jimmy Ba. 2014.
\newblock \href{http://arxiv.org/abs/1412.6980}{Adam: {A} method for stochastic
  optimization}.
\newblock {\em CoRR\/} abs/1412.6980.
\newblock
  \href{http://arxiv.org/abs/1412.6980}{http://arxiv.org/abs/1412.6980}.

\bibitem[{Lin et~al.(2017)Lin, Feng, dos Santos, Yu, Xiang, Zhou, and
  Bengio}]{DBLP:journals/corr/LinFSYXZB17}
Zhouhan Lin, Minwei Feng, C{\'{\i}}cero~Nogueira dos Santos, Mo~Yu, Bing Xiang,
  Bowen Zhou, and Yoshua Bengio. 2017.
\newblock \href{http://arxiv.org/abs/1703.03130}{A structured self-attentive
  sentence embedding}.
\newblock {\em CoRR\/} abs/1703.03130.
\newblock
  \href{http://arxiv.org/abs/1703.03130}{http://arxiv.org/abs/1703.03130}.

\bibitem[{Liu et~al.(2016)Liu, Sun, Lin, and
  Wang}]{DBLP:journals/corr/LiuSLW16}
Yang Liu, Chengjie Sun, Lei Lin, and Xiaolong Wang. 2016.
\newblock \href{http://arxiv.org/abs/1605.09090}{Learning natural language
  inference using bidirectional {LSTM} model and inner-attention}.
\newblock {\em CoRR\/} abs/1605.09090.
\newblock
  \href{http://arxiv.org/abs/1605.09090}{http://arxiv.org/abs/1605.09090}.

\bibitem[{MacCartney(2009)}]{MacCartneyThesis}
Bill MacCartney. 2009.
\newblock {\em Natural Language Inference\/}.
\newblock Ph.D. thesis, Stanford University.

\bibitem[{Mou et~al.(2016)Mou, Men, Li, Xu, Zhang, Yan, and Jin}]{Mou:P16-2022}
Lili Mou, Rui Men, Ge~Li, Yan Xu, Lu~Zhang, Rui Yan, and Zhi Jin. 2016.
\newblock \href{https://doi.org/10.18653/v1/P16-2022}{Natural language
  inference by tree-based convolution and heuristic matching}.
\newblock In {\em Proceedings of the 54th Annual Meeting of the Association for
  Computational Linguistics (Volume 2: Short Papers)\/}. Association for
  Computational Linguistics, pages 130--136.
\newblock
  \href{https://doi.org/10.18653/v1/P16-2022}{https://doi.org/10.18653/v1/P16-2022}.

\bibitem[{Munkhdalai and
  Yu(2016{\natexlab{a}})}]{DBLP:journals/corr/MunkhdalaiY16}
Tsendsuren Munkhdalai and Hong Yu. 2016{\natexlab{a}}.
\newblock \href{http://arxiv.org/abs/1607.04315}{Neural semantic encoders}.
\newblock {\em CoRR\/} abs/1607.04315.
\newblock
  \href{http://arxiv.org/abs/1607.04315}{http://arxiv.org/abs/1607.04315}.

\bibitem[{Munkhdalai and
  Yu(2016{\natexlab{b}})}]{DBLP:journals/corr/MunkhdalaiY16b}
Tsendsuren Munkhdalai and Hong Yu. 2016{\natexlab{b}}.
\newblock \href{http://arxiv.org/abs/1607.04492}{Neural tree indexers for text
  understanding}.
\newblock {\em CoRR\/} abs/1607.04492.
\newblock
  \href{http://arxiv.org/abs/1607.04492}{http://arxiv.org/abs/1607.04492}.

\bibitem[{Nangia et~al.(2017)Nangia, Williams, Lazaridou, and
  Bowman}]{nangia2017repeval}
Nikita Nangia, Adina Williams, Angeliki Lazaridou, and Samuel~R. Bowman. 2017.
\newblock The repeval 2017 shared task: Multi-genre natural language inference
  with sentence representations.
\newblock In {\em Proceedings of RepEval 2017: The Second Workshop on
  Evaluating Vector Space Representations for NLP\/}. Association for
  Computational Linguistics.

\bibitem[{Pennington et~al.(2014)Pennington, Socher, and
  Manning}]{Pennington:D14-1162}
Jeffrey Pennington, Richard Socher, and Christopher Manning. 2014.
\newblock \href{https://doi.org/10.3115/v1/D14-1162}{Glove: Global vectors for
  word representation}.
\newblock In {\em Proceedings of the 2014 Conference on Empirical Methods in
  Natural Language Processing (EMNLP)\/}. Association for Computational
  Linguistics, pages 1532--1543.
\newblock
  \href{https://doi.org/10.3115/v1/D14-1162}{https://doi.org/10.3115/v1/D14-1162}.

\bibitem[{Vendrov et~al.(2015)Vendrov, Kiros, Fidler, and
  Urtasun}]{DBLP:journals/corr/VendrovKFU15}
Ivan Vendrov, Ryan Kiros, Sanja Fidler, and Raquel Urtasun. 2015.
\newblock \href{http://arxiv.org/abs/1511.06361}{Order-embeddings of images and
  language}.
\newblock {\em CoRR\/} abs/1511.06361.
\newblock
  \href{http://arxiv.org/abs/1511.06361}{http://arxiv.org/abs/1511.06361}.

\bibitem[{Williams et~al.(2017)Williams, Nangia, and
  Bowman}]{DBLP:journals/corr/WilliamsNB17}
Adina Williams, Nikita Nangia, and Samuel~R. Bowman. 2017.
\newblock \href{http://arxiv.org/abs/1704.05426}{A broad-coverage challenge
  corpus for sentence understanding through inference}.
\newblock {\em CoRR\/} abs/1704.05426.
\newblock
  \href{http://arxiv.org/abs/1704.05426}{http://arxiv.org/abs/1704.05426}.

\bibitem[{Zhang et~al.(2017)Zhang, Zhu, Chen, Dai, Wei, and
  Jiang}]{Zhang:qa:2017}
Junbei Zhang, Xiaodan Zhu, Qian Chen, Lirong Dai, Si~Wei, and Hui Jiang. 2017.
\newblock \href{https://arxiv.org/abs/1703.04617}{Exploring question
  understanding and adaptation in neural-network-based question answering}.
\newblock {\em CoRR\/} abs/arXiv:1703.04617v2.
\newblock
  \href{https://arxiv.org/abs/1703.04617}{https://arxiv.org/abs/1703.04617}.

\end{thebibliography}
\bibliographystyle{emnlp_natbib}

\end{document}